\documentclass[letterpaper, 10 pt, conference]{ieeeconf}  % Comment this line out if you need a4paper

\IEEEoverridecommandlockouts                              % This command is only needed if 
\usepackage{amssymb}
\usepackage{amsmath}
\usepackage{amsfonts}       % blackboard math symbols
\usepackage{booktabs}       % professional-quality tables
\usepackage{balance}
\usepackage[utf8]{inputenc} % allow utf-8 input
\usepackage[T1]{fontenc}    % use 8-bit T1 fonts
\usepackage[pdftex, pdfstartview={FitV}, pdfpagelayout={TwoColumnLeft},bookmarksopen=true,plainpages = false, colorlinks=true, linkcolor=black, citecolor = black, urlcolor = black,filecolor=black , pagebackref=false,hypertexnames=false, plainpages=false, pdfpagelabels,bookmarks=false ]{hyperref}
\usepackage{url}            % simple URL typesetting
\usepackage{nicefrac}       % compact symbols for 1/2, etc.
\usepackage{microtype}      % microtypography
\usepackage{subcaption}
\usepackage{graphicx}
\usepackage{epstopdf} % needs to be AFTER graphicxs
\usepackage{mathtools}
\usepackage{marginnote}
\usepackage[dvipsnames]{xcolor}
\usepackage{float}
\usepackage[capitalize]{cleveref}
\usepackage[]{algorithmic}
\usepackage{makecell}
\usepackage{multirow}
\usepackage{paralist}
\floatstyle{boxed} \floatname{algorithm}{Algorithm}
\newfloat{algorithm}{t}{loa}[section]
\usepackage{xcolor}
\usepackage[sort,compress]{cite}
\usepackage{soul}

\usepackage{color}
\usepackage{soul}
\usepackage{xspace}
\usepackage[capitalize]{cleveref}
\usepackage{dsfont}
\usepackage{bm}
\usepackage{bbm}
\usepackage{nicefrac}  % compact symbols for 1/2, etc.
\usepackage{amsfonts}
\usepackage{amssymb}
\usepackage{array}
\usepackage{mathtools}
\usepackage{fontawesome}
\usepackage{thmtools}
\usepackage{thm-restate}
\usepackage{lipsum}
\usepackage{pifont}
\usepackage{makecell}
\usepackage{multirow}
\usepackage{afterpage}
\usepackage{tabularx,balance} 
\usepackage{url}
\usepackage[normalem]{ulem}
\usepackage{wrapfig}  % For wrapping figure
\usepackage{textcomp}

\usepackage{caption}
\usepackage{subcaption}
\DeclareCaptionLabelSeparator{periodspace}{.\quad}
\DeclareMathOperator*{\argmin}{argmin}

\DeclarePairedDelimiter{\Verts}{\lVert}{\rVert}
\NewDocumentCommand{\norm}{}{\Verts}

\title{\LARGE \bf
Efficient Deep Learning of Robust, Adaptive Policies \\ using  Tube MPC-Guided Data Augmentation}

\author{Tong Zhao$^{1,*}$, Andrea Tagliabue$^{2,*}$, Jonathan P. How$^{2}$% <-this % stops a space
\thanks{*Equal contribution}% <-this % stops a space
\thanks{$^{1}$ Department of Electrical Engineering and Computer Science, Massachusetts Institute of Technology. \texttt{tzhao@mit.edu}}%
\thanks{$^{2}$ Department of Aeronautics and Astronautics, Massachusetts Institute of Technology. \tt\{atagliab, jhow\}@mit.edu}%
\thanks{This work was funded by the Air Force Office of Scientific Research MURI FA9550-19-1-0386.}
}

\usepackage[nolist,nohyperlinks]{acronym}
\begin{acronym}
\acro{OC}{Optimal Control}
\acro{LQR}{Linear Quadratic Regulator}
\acro{MAV}{Micro Aerial Vehicle}
\acro{GPS}{Guided Policy Search}
\acro{UAV}{Unmanned Aerial Vehicle}
\acro{MPC}{model predictive control}
\acro{RTMPC}{robust tube model predictive controller}
\acro{DNN}{deep neural network}
\acro{BC}{Behavior Cloning}
\acro{DR}{Domain Randomization}
\acro{SA}{Sampling Augmentation}
\acro{SAMA}{Sampling Augmentation for Motion Adaptation}
\acro{IL}{Imitation Learning}
\acro{DAgger}{Dataset-Aggregation}
\acro{MDP}{Markov Decision Process}
\acro{VIO}{Visual-Inertial Odometry}
\acro{VSA}{Visuomotor Sampling Augmentation}
\acro{CoM}{Center of Mass}
\acro{SD}{Standard Deviation}
\acro{MSE}{Mean Squared Error}
\acro{RMSE}{Root Mean Square Error}
\acro{CSWaP}{cost, size, weight and power}
\newacro{MAV}{Micro Aerial Vehicle}
\newacro{UKF}{Unscented Kalman filter}
\newacro{EKF}{extended Kalman filter}
\newacro{PF}{particle filter}
\newacro{LSTM}{Long Short-Term Memory}
\newacro{RNN}{recurrent neural network}
\newacro{CNN}{convolutional neural network}
\newacro{CoM}{center of mass}
\newacro{USQUE}{Unscented Quaternion estimator}
\newacro{UT}{Unscented Transformation}
\newacro{MSE}{Mean Squared Error}
\newacro{GP}{Gaussian Process}
\acrodefplural{GP}[GPs]{Gaussian Processes}
\newacro{MAP}{Maximum a Posterior}
\newacro{IMU}{Inertial Measurement Unit}
\newacro{RMSE}{Root-Mean-Square error}
\newacro{RTE}{Relative Translation error}
\newacro{AIO}{Airflow-Inertial Odometry}
\newacro{VO}{visual odometry}
\newacro{RBF}{radial basis function}
\acro{OC}{Optimal Control}
\acro{LQR}{Linear Quadratic Regulator}
\acro{MAV}{Micro Aerial Vehicle}
\acro{GPS}{Guided Policy Search}
\acro{UAV}{Unmanned Aerial Vehicle}
\acro{MPC}{model predictive control}
\acro{RTMPC}{robust tube model predictive controller}
\acro{DNN}{deep neural network}
\acro{BC}{Behavior Cloning}
\acro{DR}{Domain Randomization}
\acro{SA}{Sampling Augmentation}
\acro{IL}{Imitation Learning}
\acro{DAgger}{Dataset-Aggregation}
\acro{MDP}{Markov Decision Process}
\acro{VIO}{Visual-Inertial Odometry}
\acro{VSA}{Visuomotor Sampling Augmentation}
\acro{CoM}{Center of Mass}
\acro{SD}{Standard Deviation}
\acro{MSE}{Mean Squared Error}
\acro{RMSE}{Root Mean Square Error}
\acro{NeRF}{Neural Radiance Field}
\acro{RMA}{Rapid Motor Adaptation}
\acro{RL}{Reinforcement Learning}
\acro{MRAC}{Model Reference Adaptive Control}
\acro{MLP}{Multi-Layer Perceptron}
\acro{DO}{disturbance observer}
\end{acronym}

\begin{document}

\maketitle
% Page numbers for arxiv
\thispagestyle{plain} % \thispagestyle{empty}
\pagestyle{plain} % \pagestyle{empty}

%%%%%%%%%%%%%%%%%%%%%%%%%%%%%%%%%%%%%%%%%%%%%%%%%%%%%%%%%%%%%%%%%%%%%%%%%%%%%%%%
\begin{abstract}
The deployment of agile autonomous systems in challenging, unstructured environments requires adaptation capabilities and robustness to uncertainties. Existing robust and adaptive controllers, such as those based on model predictive control (MPC), can achieve impressive performance at the cost of heavy online onboard computations. Strategies that efficiently learn robust and onboard-deployable policies from MPC have emerged, but they still lack fundamental adaptation capabilities. In this work, we extend an existing efficient Imitation Learning (IL) algorithm for robust policy learning from MPC with the ability to learn policies that adapt to challenging model/environment uncertainties. The key idea of our approach consists in modifying the IL procedure by conditioning the policy on a learned lower-dimensional model/environment representation that can be efficiently estimated online. We tailor our approach to the task of learning an adaptive position and attitude control policy to track trajectories under challenging disturbances on a multirotor. Evaluations in simulation show that a high-quality adaptive policy can be obtained in about $1.3$ hours. We additionally empirically demonstrate rapid adaptation to in- and out-of-training-distribution uncertainties, achieving a $6.1$ cm average position error under wind disturbances that correspond to about $50\%$ of the weight of the robot, and that are $36\%$ larger than the maximum wind seen during training.
\end{abstract}

%%%%%%%%%%%%%%%%%%%%%%%%%%%%%%%%%%%%%%%%%%%%%%%%%%%%%%%%%%%%%%%%%%%%%%%%%%%%%%%%
\section{Introduction} \label{sec:introduction}
The deployment of agile robots in uncertain environments demands strong robustness and rapid onboard adaptation capabilities. Approaches based on \ac{MPC}, such as robust/adaptive \ac{MPC} \cite{lopez2019adaptive, how2021performance, bujarbaruah2018adaptive, hanover2021performance, saviolo2022active}, achieve impressive robustness and adaptation performance under real-world uncertainties, but their computational cost, associated with solving a large optimization problem online, hinders deployment on computationally constrained platforms.

Recent strategies \cite{carius2020mpc, reske2021imitation, kaufmann2020deep, tagliabue2022demonstration, tagliabue2022robust, tagliabue2022output} avoid solving onboard the large optimization problem associated with MPC by instead deploying a fast \ac{DNN} policy, obtained from \ac{MPC} via \ac{IL}. These procedures consider the \ac{MPC} as an ``expert'' which provides task-relevant demonstrations used to train a ``student'' policy, using \ac{IL} algorithms such as \ac{BC} \cite{argall2009survey} or \ac{DAgger} \cite{ross2011reduction}.
Among these methods, our previous work, called \ac{SA} \cite{tagliabue2022demonstration}, enables the training of \ac{DNN} policies that achieve high robustness to uncertainties in a demonstration-efficient manner. \ac{SA} leverages a specific type of robust \ac{MPC}, called \ac{RTMPC}, to i) collect demonstrations that account for the effects of uncertainties, and ii) efficiently generate extra data that robustifies and makes more demonstration-efficient the learning procedure. However, while \ac{SA} can efficiently train \textit{robust} \ac{DNN} policies capable of controlling a variety of robots \cite{tagliabue2022output, tagliabue2022demonstration}, the obtained policies cannot adapt to the effects of model and environment uncertainties, resulting in large errors when subject to large disturbances.

\begin{figure}
    \centering
    \includegraphics[trim= 12 12 7 0,clip,width=0.45\textwidth]{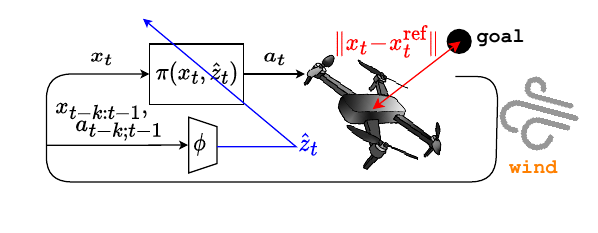}\linebreak[0]%
    \includegraphics[trim=0 0 0 0,clip,width=0.48\textwidth]{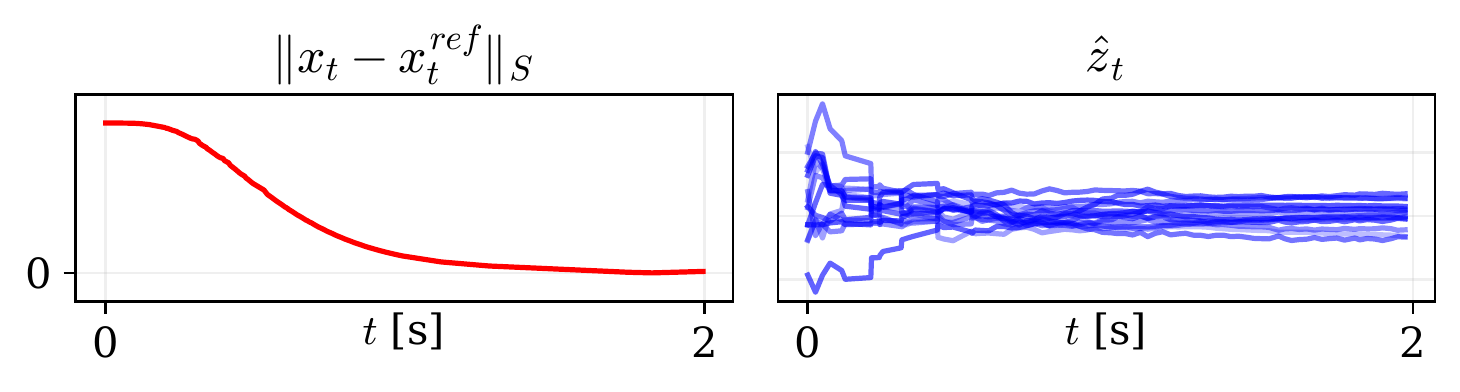}\linebreak[0]%

    \vspace{-1.5ex}
    \caption{Diagram of our combined \textit{position and attitude} controller, efficiently learned from Robust Tube MPC using Imitation Learning. The \textit{adaptation module} $\phi$ uses a history of $k$ states $x_{t-k:t-1}$ and actions $a_{t-k:t-1}$ to estimate the \textit{extrinsics} $\hat{z}_t$, a low dimensional representation of the environment parameters. This estimate allows the \textit{base policy} (controller) $\pi$ to adapt to various changes in the system, reducing error in the face of model/environment uncertainties. Selection matrix $S$ selects the position axis along the direction of the wind.}
    \label{fig:cover-img}
    \vspace{-2em}
\end{figure}

In this work we present \ac{SAMA}, a strategy to efficiently generate a \textit{robust and adaptive} \ac{DNN} policy using \ac{RTMPC}, enabling the policy to compensate for large uncertainties and disturbances. The key idea of our approach, summarized in \cref{fig:cover-img} and \cref{fig:approach_diagram}, is to augment the efficient \ac{IL} strategy \ac{SA}\cite{tagliabue2022demonstration} with an adaptation scheme inspired by the recently proposed adaptive policy learning method \ac{RMA} \cite{kumar2021rma}.  \ac{RMA} uses \ac{RL} to train in simulation a fast \ac{DNN} policy whose inputs include a learned low-dimensional representation of the model/environment parameters experienced during training. At deployment, this low-dimensional representation is estimated online, triggering adaptation. \ac{RMA} policies have demonstrated impressive adaptation and generalization performance on a variety of robots/conditions \cite{kumar2021rma, kumar2022adapting, zhang2022zero, qi2022hand}. 
Similar to \ac{RMA}, in our work we include to the inputs of the learned policy a low-dimensional representation of model/environment parameters that could be encountered at deployment. These parameters are experienced during demonstration collection from \ac{MPC} and can be efficiently estimated online, enabling adaptation. Unlike \ac{RMA}, however, we bypass the challenges associated with \ac{RL}, such as reward selection and tuning, via an efficient \ac{IL} procedure using our \ac{MPC}-guided data augmentation strategy \cite{tagliabue2022demonstration}. 
We tailor our approach to the challenging task of \textit{trajectory tracking} for a multirotor, designing a policy that controls both \textit{position and attitude} of the robot and that is capable of adapting to uncertainties in the translational and rotational dynamics. 
Our evaluation, performed under challenging model errors and disturbances in a simulation environment, demonstrates rapid adaptation to in- and out-of-distribution uncertainties while tracking agile trajectories with top speeds of $3.2$ m/s, using an adaptive policy that is learned in $1.3$ hours. This differs from prior \ac{RMA} work for quadrotors \cite{zhang2022zero}, where the focus of adaptation is only on \textit{attitude} control during quasi-static trajectories.
Additionally, \ac{SAMA} shows comparable performance to \ac{RTMPC} combined with a high-performance, state-of-the-art but significantly more computationally expensive \ac{DO}.

\noindent
\textbf{Contributions.}
\begin{inparaenum}
\item We augment our previous efficient and robust \ac{IL} strategy from \ac{MPC} \cite{tagliabue2022demonstration} with the ability to learn robust \textit{and adaptive} policies, therefore reducing tracking errors under uncertainties while maintaining high learning efficiency. Key to our work is to leverage the performance of an RMA-like \cite{kumar2021rma} adaptation scheme, but without relying on \ac{RL}, therefore avoiding reward selection and tuning.
\item We apply our proposed methodology to the task of adaptive position and attitude control for a multirotor, demonstrating for the first time \ac{RMA}-like adaptation to uncertainties that cause position and orientation errors, unlike previous work \cite{zhang2022zero} that only focuses on adaptive attitude control. 
\end{inparaenum}

\section{Related Works} \label{sec:relaed_works}
\noindent
\textbf{Adaptive Control.}
Adaptation strategies can be classified into two categories, direct and indirect methods. Indirect methods aim at explicitly estimating models or parameters, and these estimates are leveraged in model-based controllers, such as \ac{MPC} \cite{borrelli2017predictive}. Model/parameter identification include filtering techniques \cite{svacha2020imu, wuest2019online}, disturbance observers \cite{tagliabue2020touch, tagliabue2019robust, mckinnon2016unscented}, set-membership identification methods \cite{lopez2019adaptive, how2021performance} or active, learning-based methods \cite{saviolo2022active}. While these approaches achieve impressive performance, they often suffer from high computational cost due to the need of identifying model parameters online and, when \ac{MPC}-based strategies are considered, solving large optimization problems online.
Direct methods, instead, develop policy updates that improve a certain performance metric. This metric is often based on a reference model, while the updates involve the shallow layers of the \ac{DNN} policy \cite{joshi2019deep, joshi2020design, zhou2021bridging}. Additionally, policy update strategies can be learned offline using meta-learning \cite{richards2021adaptive, oconnel2022neural}. While these methods employ computationally-efficient \ac{DNN} policies, they require extra onboard computation to update the policy, require costly offline training procedures, and/or do not account for actuation constraints. 
Parametric adaptation laws, such as $\mathcal{L}_1$ adaptive control \cite{hovakimyan20111}, have been applied to the control inputs generated by \ac{MPC} \cite{pravitra2020} \cite{hanover2021performance}, significantly improving \ac{MPC} performance; however, these approaches still require solving onboard the large optimization problem associated with \ac{MPC}, and do not account for control limits.
Our work leverages the inference speed of a \ac{DNN} for computationally-efficient onboard deployment, training the policy using an efficient \ac{IL} procedure (our previous work \cite{tagliabue2022demonstration}) that uses a robust \ac{MPC} capable of accounting for state and actuation constraints. 

\noindent
\textbf{\acf{RMA}.}
\ac{RMA} \cite{kumar2021rma} has recently emerged as a high-performance, hybrid adaptive strategy. Its key idea is to learn a \ac{DNN} policy conditioned on a low-dimensional (encoded) model/environment representation that can be efficiently inferred online using another \ac{DNN}. The policy is trained using \ac{RL}, in a simulation where it experiences different instances of the model uncertainties/disturbances. \ac{RMA} policies have controlled a wide range of robots, including quadruped \cite{kumar2021rma}, biped \cite{kumar2022adapting}, hand-like manipulators \cite{qi2022hand} and multirotors \cite{zhang2022zero}, demonstrating rapid adaptation and generalization to previously unseen disturbances. Our work takes inspiration from the adaptation strategy introduced by \ac{RMA}, as we learn policies conditioned on a low-dimensional environment representation that is estimated online. However, unlike \ac{RMA}, our learning procedure does not require the reward selection and tuning typically encountered in \ac{RL}, as it leverages an efficient \ac{IL} strategy from robust \ac{MPC}. An additional difference to \cite{zhang2022zero}, where \ac{RMA} is used to generate a policy for attitude control of multirotors of different sizes, is that our work focuses on the challenging task of learning an adaptive trajectory tracking controller, which compensates for the effects of uncertainties on its attitude \textit{and} position.

\noindent
\textbf{Efficient Imitation Learning from MPC.}
Our previous work \cite{tagliabue2022demonstration} has focused on designing efficient policy learning procedures by leveraging \ac{IL} and a \ac{RTMPC} to guide demonstration collection and co-design a demonstration-efficient data augmentation procedure. This approach has enabled efficient learning of policies capable of performing trajectory tracking on multirotors \cite{tagliabue2022demonstration} and sub-gram, soft-actuated flapping-wings aerial vehicles \cite{tagliabue2022robust}, also enabling efficient sensorimotor policy learning \cite{tagliabue2022output}. In this work, we extend this procedure by enabling online \textit{adaptation}, while maintaining high efficiency (in terms of demonstrations and training time) of the policy learning procedure.

\section{Preliminaries} \label{sec:preliminaries}
Our approach leverages and modifies two key algorithms, \ac{RTMPC} \cite{mayne2005rtmpc} and \ac{RMA} \cite{kumar2021rma}. These methods are summarized in the following parts for completeness. 

\noindent
\textbf{Notation:}
Consider two sets $\mathbb{X} \subseteq \mathbb{R}^{n}$ and $\mathbb{Y} \subseteq \mathbb{R}^{n}$, and a linear map $M \in \mathbb{R}^{m \times n}$. We define the sets: 
\begin{itemize}
\item $M\mathbb{X} :=\{Mx \mid x \in \mathbb{X}\}$; 
\item $\mathbb{X} \oplus \mathbb{Y}\!:=\!\{x+y\mid\!x\!\in\!\mathbb{X}, y\!\in\!\mathbb{Y}\}$ (Minkowski sum);
\item $\mathbb{X} \ominus \mathbb{Y}\!:=\!\{z\mid\!z+y\!\in\!\mathbb X, \forall y\!\in\!\mathbb Y\}$ (Pontryagin difference). 
\end{itemize}

\subsection{Robust Tube MPC (RTMPC)}
\label{sec:rtmpc_design}
\ac{RTMPC} \cite{mayne2005rtmpc} is a type of robust \ac{MPC}, capable of regulating a linear system assumed subject to bounded uncertainty, while guaranteeing state and actuation constraint satisfaction. The linearized, discrete-time model of the system is given by
\begin{equation}
\label{eqn:rtmpc-model}
    x_{t+1} = Ax_t + Bu_t + w_t,
\end{equation}
where $x \in \mathbb{R}^{n_x}$ is the state (size $n_x$), $u \in \mathbb{R}^{n_u}$ are the commanded actions (size $n_u$), and $w \in \mathbb{W} \subseteq \mathbb{R}^{n_x}$ is an additive state disturbance. Matrices $A \in \mathbb{R}^{n_x \times n_x}$ and $B \in \mathbb{R}^{n_x \times n_u}$ are the nominal robot dynamics. The system is additionally subject to state and actuation constraints $x\!\in\!\mathbb{X}$, $u\!\in\!\mathbb{U}$, and the disturbance/uncertainty is assumed to be bounded, i.e., $w\!\in\!\mathbb{W}$. At every discrete timestep $t$, an estimate of the state of the actual system $x_t$ and a reference trajectory $\mathbf{x}^{\text{ref}}_t := [x^{\text{ref}}_{0,t}, \hdots, x^{\text{ref}}_{N,t}]$ are provided. The controller then plans a sequence of states $\check{\mathbf{x}}_t := [\check{x}_{0,t}, \hdots, \check{x}_{N,t}]$ and actions $\check{\mathbf{u}}_t := [\check{u}_{0,t}, \hdots, \check{u}_{N-1,t}]$ that solve the optimization problem: 
\begin{equation}
\label{eqn:rtmpc-optimization-constrained}
\begin{gathered}[b]
    (\check{\mathbf{x}}^*_t, \check{\mathbf{u}}^*_t) = \argmin_{\check{\mathbf{x}}_t, \check{\mathbf{u}}_t}
    V(\check{\mathbf{x}}_t, \check{\mathbf{u}}_t, \mathbf{x}_t^{\text{ref}}, x_t)
    \\
    \text{subj. to } 
    \begin{lgathered}[t]
        \check{x}_{n+1,t} = A \check{x}_{n,t} + B \check{u}_{n,t},
        \\
        \check{x}_{n,t} \in \mathbb{X}_c , \: \check{u}_{n,t} \in \mathbb{U}_c , \: x_t \in  \check{x}_{0,t} \oplus \mathbb{Z},
    \end{lgathered}
    \\
\end{gathered}
\end{equation}
with $n = 0, \dots, N-1$. The objective function is:
\begin{equation}
\label{eqn:rtmpc-objective}
\begin{gathered}[b]
    V(\check{\mathbf{x}}_t, \check{\mathbf{u}}_t, \mathbf{x}^{\text{ref}}_t, x_t) := 
    \norm{e_{N,t}}_P^2 \!+\!\! \sum_{n=0}^{N-1} \norm{e_{n,t}}_Q^2 \!+\! \norm{\check{u}_{n,t}}_R^2,
\end{gathered}
\end{equation}
where $e_{n,t} := \check{x}_{n, t} - x^{\text{ref}}_{n, t}$, and $\check{\cdot}$ denotes internal variables of the optimization. Positive definite matrices $Q \in \mathbb{R}^{n_x \times n_x}$ and $R \in \mathbb{R}^{n_u \times n_u}$ are user-selected weights that define the stage cost, while $P \in \mathbb{R}^{n_x \times n_x}$ is a positive definite matrix that represents the terminal cost. Dynamic feasibility is enforced by the constraints $\check{x}_{n+1,t} = A \check{x}_{n,t} + B \check{u}_{n,t}$. The $\check{\mathbf{x}}_t$ and $\check{\mathbf{u}}_t$ obtained by minimizing the objective in \cref{eqn:rtmpc-objective} are denoted $\check{\mathbf{x}}_t^* := [\check{x}_{0,t}^*, \hdots, \check{x}_{N,t}^*]$ and $\check{\mathbf{u}}_t^* := [\check{u}_{0,t}^*, \hdots, \check{u}_{N-1,t}^*]$.

Given $\check{\mathbf{x}}_t^*$ and $\check{\mathbf{u}}_t^*$, the control input is then produced by a feedback policy, called an \textit{ancillary controller}
\begin{equation}
\label{eqn:rtmpc-ancillary}
    u_t = \check{u}_{0,t}^* + K (x_t - \check{x}_{0,t}^*),
\end{equation}
where $K$ is a feedback gain matrix, chosen such that the matrix $A_K := A+BK$ is Schur stable, for example, by solving the infinite horizon, discrete-time LQR problem using $(A, B, Q, R)$. Given $K$, we compute a \textit{disturbance invariant set} $\mathbb{Z} \subseteq \mathbb{R}^{n_x}$, the cross section of a \textit{tube}, which satisfies $A_K \mathbb{Z} \oplus \mathbb{W} \subseteq \mathbb{Z}$.
The ancillary controller with gain $K$ ensures that, in the controlled system, if $x_t \in \check{x}_{0,t}^* \oplus \mathbb{Z}$, then $x_{t+1} \in \check{x}_{0, t+1}^* \oplus \mathbb{Z}$; i.e., $x_{t+1}$ remains in the tube centered around $\check{x}_{0,t+1}$ for all realizations of  $w_t \in \mathbb{W}$ \cite{mayne2001robust}.  

To account for the output of the ancillary controller, \ac{RTMPC} tightens state and actuation constraints in the optimization (\ref{eqn:rtmpc-objective}) according to $\mathbb{X}_c := \mathbb{X} \ominus \mathbb{Z}$, $\mathbb{U}_c := \mathbb{U} \ominus K\mathbb{Z}$.
Additionally, the controller constrains $\check{x}_{0,t}$ so that, at time $t$, the current state $x_t$ is within a tube centered around $\check{x}_{0,t}$. These restrictions combined ensure that the ancillary controller keeps the actual state $x$ in a tube centered around optimal planned state $\check{x}_0^*$, while simultaneously ensuring that the combined controller remains within state and actuation constraints for any realization of $w \in \mathbb{W}$. 

\subsection{Rapid Motor Adaptation (RMA)}

\ac{RMA} \cite{kumar2021rma} enables learning of adaptive control policies in simulation using model-free \ac{RL}.  The key idea of RMA is to learn a policy composed of a base policy $\pi$ and an adaptation module $\phi$. The base policy is denoted as:
\begin{equation}
a_t = \pi(x_t, z_t),
\end{equation}
and takes as input the current state $x_t \in \mathbb{R}^{n_x}$ and an \textit{extrinsics} vector $z_t \in \mathbb{R}^{n_z}$, outputting commanded actions $a_t \in \mathbb{R}^{n_a}$. Key to this method is the extrinsics vector $z_t$, a low-dimensional representation of an environment vector $e_t \in \mathbb{R}^{n_e}$, which captures all the possible parameters/disturbances that may vary at deployment time  (i.e., mass, drag, external disturbances, \dots), and towards which the policy should be able to adapt. However, because $e_t$ is not directly accessible in the real world, it is not possible to directly compute $z_t$ at deployment time. Instead, \ac{RMA} produces an estimate $\hat{z}_t$ of $z_t$ via an \textit{adaptation module} $\phi$: 
\begin{equation}
    \hat{z}_t = \phi(x_{t-k : t-1}, a_{t-k; t-1}), 
\end{equation}
whose input is a history of the $k$ past states $x_{t-k : t-1}$ and actions $a_{t-k:t-1}$ at  deployment, enabling rapid adaptation.  

Learning $\phi$ and $\pi$ is divided in two \textit{phases}.
\subsubsection{Phase 1: Base Policy and Environment Factor Encoder}
In \textit{Phase 1}, \ac{RMA} trains the base policy $\pi$ and an intermediate policy, the environment factor encoder $\mu$:
\begin{equation}
\label{eq:env_factor_encoder}
z_t = \mu(e_t)
\end{equation}
which takes as input the vector $e_t$ and produces the \textit{extrinsics} vector $z_t$. 
The two modules are trained using model-free \ac{RL} (e.g., PPO \cite{schulman2017proximal}) in a simulation environment subject to instances $e_t$ of the possible model/environment uncertainties, and leveraging a reward function that captures the desired control objective. Using this procedure, \ac{RL} discovers policies that can perform well under disturbances/uncertainties. 
\subsubsection{Phase 2: Adaptation Module}
\label{eqn:rma_adaptation_module}
The adaptation module $\phi$ is obtained by generating a dataset of state-action histories in simulation via
\begin{align}
    \hat{z}_t &= \phi(x_{t-k : t-1}, a_{t-k; t-1}), \\
          a_t &= \pi(x_t, \hat{z}_t).
\end{align}
Because we have access to the ground truth environment parameters $e_t$ in simulation, \ac{RMA} can  compute $z_t$ at every timestep using \cref{eq:env_factor_encoder}, allowing us to train $\phi$ via supervised regression, minimizing the \ac{MSE} loss $\norm{z_t - \hat{z}_t}^2$. This is done iteratively, by alternating the collection of on-policy rollouts with updates of $\phi$.

\section{Approach} \label{sec:approach}
\begin{figure*}
    % [trim = {left, bottom, right, top}, clip]
    \centering\includegraphics[width=0.9\textwidth,trim = {0, 20, 45, 0}, clip]{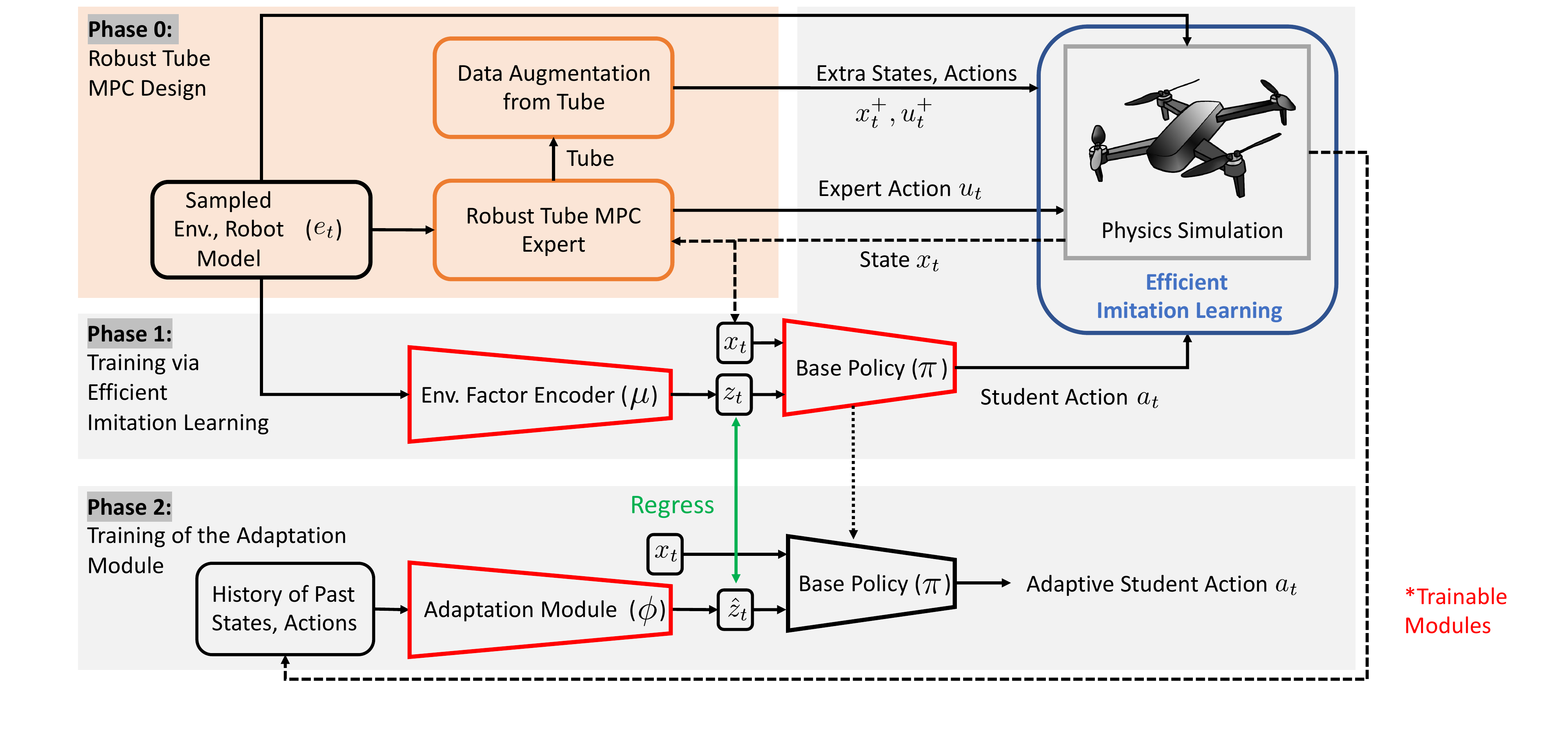}
    \vspace{-1.5ex}
    \caption{Schematic representation of \acf{SAMA}, our proposed approach for efficient learning of adaptive polices from \ac{MPC}. The key idea of \ac{SAMA} consists in leveraging an efficient Imitation Learning strategy, \acf{SA} \cite{tagliabue2022demonstration}, to collect demonstrations and perform data augmentation using a Robust Tube \ac{MPC}. This efficiently generated data is used to train a student policy conditioned on a latent representation $z_t$ of environment and robot parameters $e_t$. Following the \acf{RMA} \cite{kumar2021rma} procedure, we then train an adaptation module that can produce an estimate $\hat{z}_t$ of these environment parameters from a sequence of past states and actions. This approach enables efficient learning of a robust, adaptive policy from \ac{MPC} without leveraging \ac{RL}, avoiding any reward tuning and making use of available priors on the model of the robot.}
    \label{fig:approach_diagram}
    \vskip-3ex
\end{figure*}

The proposed approach, summarized in \cref{fig:approach_diagram}, consists in a \textit{three phase} policy learning procedure: 

\subsection{Phase 0: Robust Tube MPC Design}
\label{eqn:adaptive-rtmp}
As in \ac{RMA}, we train our policies in a simulation environment implementing the full nonlinear dynamic model of the robot/environment, with parameters (model/environment uncertainties, disturbances...) captured by the environment parameter vector $e$. At each timestep $t$, each entry in $e$ may change with some probability $p$, with entries changing independently of each other (see Table \ref{table:env-params} for more details on the distributions of $e$ in the train and test environments). Whenever $e$ changes, we update the \ac{RTMPC}, described in \cref{sec:rtmpc_design}, as follows. 
First, since the controller uses linear system dynamics, for a given environment parameter vector $e_t$ at time $t$ we compute a discrete-time linear system by discretizing and linearizing the full nonlinear system dynamics, obtaining:  
\begin{equation}
\label{eqn:rtmpc-time-varying-model}
    x_{t+1} = A(e_t) x_t + B(e_t) u_t.
\end{equation}
The linearization is performed by assuming a given desired operating point; for our multirotor-based evaluation, this point corresponds to the hover condition.

Second, the feedback gain $K_t$ for the ancillary controller in \cref{eqn:rtmpc-ancillary} is updated by solving the infinite horizon, discrete-time LQR problem using $(A(e_t), B(e_t), Q, R)$, leaving the tuning weights $Q$, $R$ fixed. Last, we compute the robust control invariant set $\mathbb{Z}_t$ employed by \ac{RTMPC} from the resulting $K_t$, $A(e_t)$, $B(e_t)$, and a given $\mathbb W$. Due to the computational cost of \textit{precisely} computing $\mathbb{Z}_t$ (from $K_t$, $A(e_t)$, $B(e_t)$, and $\mathbb{W}$), we generate an outer-approximation of $\mathbb{Z}_t$ via Monte Carlo simulation. This is done by computing the axis-aligned bounding box of the state deviations obtained by perturbing the closed loop system $A_{K_t}$ with randomly sampled instances of $w \in \mathbb{W}$. The set $\mathbb W$ is designed to capture the effects of linearization and discretization errors, as well as errors that are introduced by the learning/parameter estimation procedure. 

\subsection{Phase 1: Base Policy and Environment Factor Encoder Learning via Efficient Imitation}
We now describe the procedure to efficiently learn a base policy $\pi$ and an environment factor encoder $\mu$ in simulation. Similar to \ac{RMA}, our base policy takes as input the current state $x_t$, an extrinsics vector $z_t$ and, different from RMA, a reference trajectory $\mathbf{x}_t^{\text{ref}}$. It outputs a vector of actuator commands $a_t$. 
As in \ac{RMA}, the extrinsics vector $z_t$ represents a low dimensional encoding of the environment parameters $e_t$, and it is generated in this phase by the environment factor encoder $\mu$:
\begin{equation}
    \label{eqn:adaptive_policy}
    \begin{split}
        z_t &= \mu(e_t) \\ 
        a_t &= \pi(x_t, z_t, \mathbf{x}_t^{\text{ref}}).
    \end{split}
\end{equation}
We jointly train the base policy $\pi$ and environment encoder $\mu$ end-to-end. However, unlike \ac{RMA}, we do not use \ac{RL}, but demonstrations collected from \ac{RTMPC} in combination with \ac{DAgger} \cite{ross2011reduction}, treating the \ac{RTMPC} as a \textit{expert}, and the policy in \cref{eqn:adaptive_policy} as a \textit{student}. More specifically, at every timestep, given the environment parameters vector $e_t$, the current state of the robot $x_t$, and the reference trajectory $\mathbf{x}_t^{\text{ref}}$, the expert generates a control action $u_t$ by first computing a \textit{safe} reference plan $\check{\mathbf{x}}_t^*, \check{\mathbf{u}}_t^*$, and then by using the ancillary controller in \cref{eqn:rtmpc-ancillary}. The obtained control action is applied to the simulation with a probability $\beta$, otherwise the applied control action is queried from the student (\cref{eqn:adaptive_policy}). At every timestep, we store in a dataset $\mathcal{D}$ the (input, output) pairs $(\{\mathbf{x}_t^{\text{ref}}, x_t, e_t\}, u_t)$.

\noindent
\textbf{Tube-guided Data Augmentation.} \label{sec:approach:tube_augmentation} We leverage our previous work \cite{tagliabue2022demonstration} to augment the collected demonstrations with extra data that accounts for the effects of the uncertainties in $\mathbb W$. This procedure leverages the idea that the tube $\mathbb{Z}_t$ centered around $\check{x}_{0,t}^*$, as computed by \ac{RTMPC}, represents a model of the states that the system may visit when subject to the uncertainties captured by the additive disturbances $w \in \mathbb{W}$, while the ancillary controller \cref{eqn:rtmpc-ancillary} represents an efficient way to compute control actions that ensure the system remains inside the tube. Therefore, at each timestep $t$, given the ``safe'' plan computed by the expert $\check{\mathbf{x}}_t^*, \check{\mathbf{u}}_t^*$, we compute extra state-action pairs $(x_t^+, u_t^+)$ by sampling states from inside the tube $x_t^+ \in \check{x}_{0,t}^* \oplus \mathbb{Z}_t$, and computing the corresponding robust control action $u_t^+$ using the ancillary controller:
\begin{equation}
    u_t^+ = \check{u}_{0,t}^* + K(x_t^+ - \check{x}_{0,t}^*).
\end{equation}
In this way, we obtain extra (input, output) samples $(\{\mathbf{x}_t^{\text{ref}}, x_t^+, e_t\}, u_t^+)$ that are added to  the training dataset $\mathcal{D}$. 
Last, the policy in \cref{eqn:adaptive_policy} is trained end-to-end using the dataset $\mathcal{D}$, by finding the parameters of $\pi$ and $\mu$ that minimize the following \ac{MSE} loss: $\| u_i - \pi(x_i, \mu(e_i), \mathbf{x}_i^{\text{ref}})\|_2^2$, where $i$ denotes the $i$-th datapoint in $\mathcal{D}$.  

\subsection{Phase 2: Learning the Adaptation Module}
This step is performed as in \ac{RMA} \cite{kumar2021rma}, and is described in \cref{eqn:rma_adaptation_module} of this work.

\section{Evaluation} \label{sec:evaluation}
\subsection{Evaluation Approach}
We evaluate the proposed approach in the context of trajectory tracking for a multirotor, by learning to track an $8$ s long, heart-shaped trajectory ($\heartsuit$) and an $8$ s long, eight-shaped trajectory ($\infty$) with a maximum velocity of $3.2$ m/s. All evaluation is performed on a desktop machine with Intel i9-10920X CPUs and Nvidia RTX $3090$ GPUs.

\begin{table}[t] 
\caption{Robot/environment parameter ranges during training and testing. For most parameters, the testing range is twice as wide as the training range, allowing us to evaluate our methods under large out-of-distribution model errors and disturbances. The nominal values are the average of the training ranges. At each timestep, each entry in $e_t$ changes with $p=0.001$ in the training environment, and $p=0.002$ in the test environment.}
\label{table:env-params}
\begin{tabular}{|p{2.8cm}|p{0.8cm}|p{1.6cm}|p{1.6cm}|}
\hline
\textbf{Parameters}                                                    & \textbf{Units} & \textbf{Train Range} & \textbf{Test Range} \\ \hline \hline
Mass                                                                   & kg             & [1.0, 1.6]                     & [0.8, 1.8]           \\
Inertia ($x_B, y_B$)                                                   & kg m$^2$       & [6.6, 9.8] \hfill e-3          & [4.9, 12.0] \hfill e-3       \\
Inertia ($z_B$)                                                        & kg m$^2$       & [1.0, 1.5] \hfill e-2          & [0.8, 1.8] \hfill e-2       \\
Drag (translational)                                                   & Ns/m           & [0.08, 0.12]                   & [0.06, 0.14]           \\
Drag (rotational)                                                      & Nms/rad        & [8.0, 12.0] \hfill e-5         & [6.0, 14.0] \hfill e-5       \\
Arm length                                                             & m              & [0.13, 0.20]                   & [0.10, 0.23]           \\
Ext.~force\hphantom{.}($x_W,y_W,z_W$)                                  & N              & [-2.6, 2.6]                    & [-3.2, 3.2]            \\
Ext.~torque\hphantom{.}($x_B, y_B$)                                    & Nm             & [-0.42, 0.42]                  & [-0.53, 0.53]          \\
Ext.~torque\hphantom{.}($z_B$)                                         & Nm             & [-4.2, 4.2] \hfill e-2         & [-4.2, 4.2] \hfill e-2      \\ \hline
\end{tabular}
\vskip-5ex
\end{table}

\noindent
\textbf{Simulation Details.} \label{sec:evaluation:sim} Learning and evaluation are performed in simulation, implemented by integrating a realistic nonlinear model of the dynamics of a multirotor (with $6$ motors):
\begin{align}
\label{eqn:sim_nonlinear_model}
\dot{p}=v,
\;\;\;&\; 
m\dot{v}\!=\!f_{\text{cmd}} R_B(q)z\!-\!mg_Wz\!+\!f_{\text{drag}}\!+\!f_{\text{ext}}, & \nonumber \\ 
\dot{q}=\frac{1}{2}\Omega(\omega)q,
\;\;\;&\;
J \dot{\omega}\!=\!-\omega \times J \omega\!+\!\tau_{\text{cmd}}\!+\!\tau_{\text{drag}}\!+\!\tau_{\text{ext}}. & 
\end{align}
Position and velocity $p, v \in \mathbb{R}^3$ are expressed in the world frame $W$, $q \in \text{SO}(3)$ is the attitude quaternion, $\omega$ is the angular velocity, $m$ is the mass and $J$ is the inertia matrix assumed diagonal. The total torques $\tau_\text{cmd}$ and forces $f_\text{cmd}$ produced by the propellers, as expressed in body frame $B$, are linearly mapped to the propellers' thrust via a mixer/allocation matrix (e.g., \cite{tagliabue2020touch}).  
We assume the presence of isotropic drag forces and torques $f_\text{drag} = -c_{dv}v$ and $\tau_{\text{drag}} = -c_{d\omega}\omega$, with $c_{dv} >0$, $c_{d\omega} > 0$, and the presence of external forces $f_{\text{ext}}$ and torques $\tau_{\text{ext}}$. The environment parameter vector $e_t$ has size $13$, and contains the robot/environment parameters in \cref{table:env-params}. We use the \texttt{acados} integrator \cite{Verschueren2021} to simulate these dynamics with a discretization interval of $0.002$ s.

\noindent
\textbf{\ac{RTMPC} for Trajectory Tracking on a Multirotor.} The controller has state of size $n_x = 12$, consisting of position, velocity, Euler angles, and angular velocity. It generates thrust/torque commands ($n_u = 4$) mapped to the $6$ motor forces via allocation/mixer matrix ($n_a = 6$). We use an adversarial heuristic to find a value for $\mathbb{W}$, and specifically we assume that it matches the external forces and torques used in training distribution (Table \ref{table:env-params}), as they are close to the physical actuation limits of the platform. The reference trajectory is a sequence of desired positions and velocities for the next $1$\,s, discretized with a sampling time of $0.04$ s (corresponding to a planning horizon of $N=25$, and $300$-dim vector). The controller takes into account state constraints (i.e., available 3D flight space, velocity limits, etc) and actuation limits, and is simulated to run at $500$ Hz.

\noindent
\textbf{Student Policy Architecture.} The base policy $\pi$ is a $3$-layer \ac{MLP} with $256$-dim hidden layers, which takes as input the current state $x_t \in \mathbb{R}^{12}$ and extrinsics vector $z_t \in \mathbb{R}^{8}$ and outputs motor forces $a_t \in \mathbb{R}^6$. The environment encoder $\mu$ is a $2$-layer \ac{MLP} with $128$-dim hidden layers, taking as input environment parameters $e_t \in \mathbb{R}^{13}$. The adaptation module $\phi$ projects the latest $400$ state-action pairs into $32$-dim representations using a $2$-layer \ac{MLP}. Then, a $3$-layer 1-D CNN convolves the representation across time to capture temporal correlations in the input. The input channel number, output channel number, kernel size, and stride for each layer is $[32, 32, 8, 4]$. The flattened CNN output is linearly projected to obtain $\hat{z}_t$. Like the \ac{RTMPC} expert, the student policy is simulated to run at $500$ Hz.

\subsection{Training Details and Hyperparameters} \label{training-details}
All policies are implemented in PyTorch and trained with the Adam optimizer, with learning rate $0.001$ and default parameters.

\noindent
\textbf{Phase 1.} We train $\mu$ and $\pi$ by collecting $8$ s long trajectories, with $1$ s of hovering before and after the trajectories. The expert actions are sampled at $20$ Hz, resulting in $200$ expert actions per demonstration (when no additional samples are drawn from the tube). When drawing additional samples from the tube, we do so in two different ways. The first is to uniformly sample the tube for every demonstration we collect from the expert, extracting $N_\text{samples} = \{25, 50, 100\}$ samples per timestep; these methods are denoted as SAMA-$N_\text{samples}$. In the second way, we apply data augmentation (using $N_\text{samples} = 100$ samples per timestep) only to the first collected demonstration, while we use  DAgger only (no data augmentation) for the subsequent demonstrations. This method is denoted as SAMA-100-FT (Fine-Tuning, as DAgger is used to fine-tune a good initial guess generated via data augmentation). These different procedures enable us to study trade-offs between improving robustness/performance (more samples) or the training time (fewer samples). Across our evaluations, we always set the DAgger hyperparameter $\beta$ to $1$ for the first demonstration and $0$ otherwise.

\noindent
\textbf{Phase 2.} Similar to previous RMA-like approaches \cite{kumar2021rma,kumar2022adapting,zhang2022zero,qi2022hand}, we train $\phi$ via supervised regression.

\begin{figure}[ht!]
    \centering\includegraphics[width=0.5\textwidth,trim = {0.2cm, 0, 0.05cm, 0}, clip]{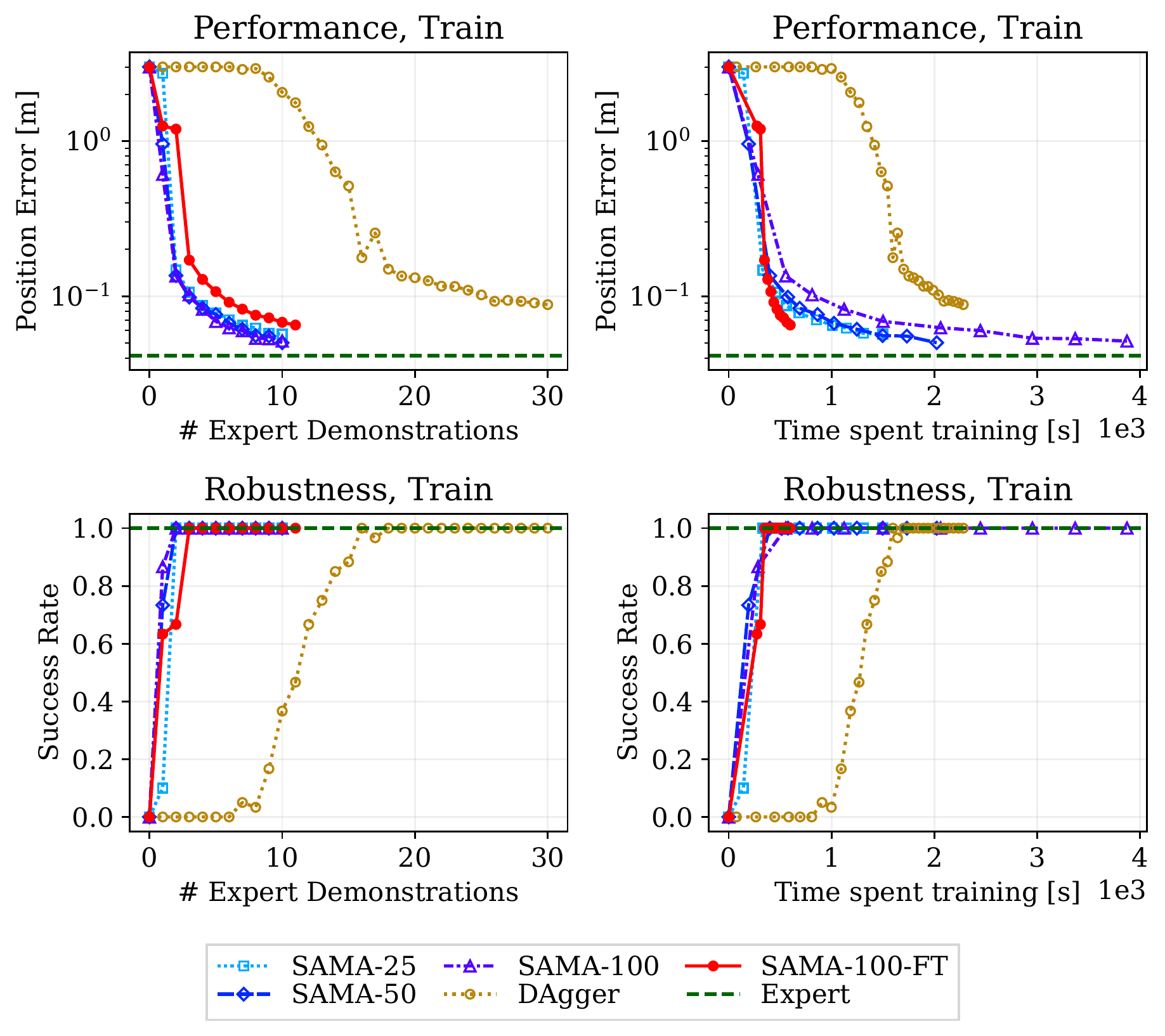}
    \captionof{figure}{Performance and robustness in the \textbf{training environment} after \textit{Phase 1}, as a function of number of demonstrations and training time. \ac{IL} allows for the learning of effective \textit{Phase 1} policies in one hour on a single core, as opposed to \ac{RL} which has been reported to take two hours on an entire desktop machine \cite{zhang2022zero}. This training time can be significantly shortened by using tube-guided data augmentation during training.}
    \label{fig:perf_robust_train}
    \vskip-2ex
\end{figure}

\subsection{Efficiency, Robustness, and Performance} 

\begin{figure}
    % trim={<left> <lower> <right> <upper>}
    \centering\includegraphics[trim=35 65 400 148,clip,width=\columnwidth]{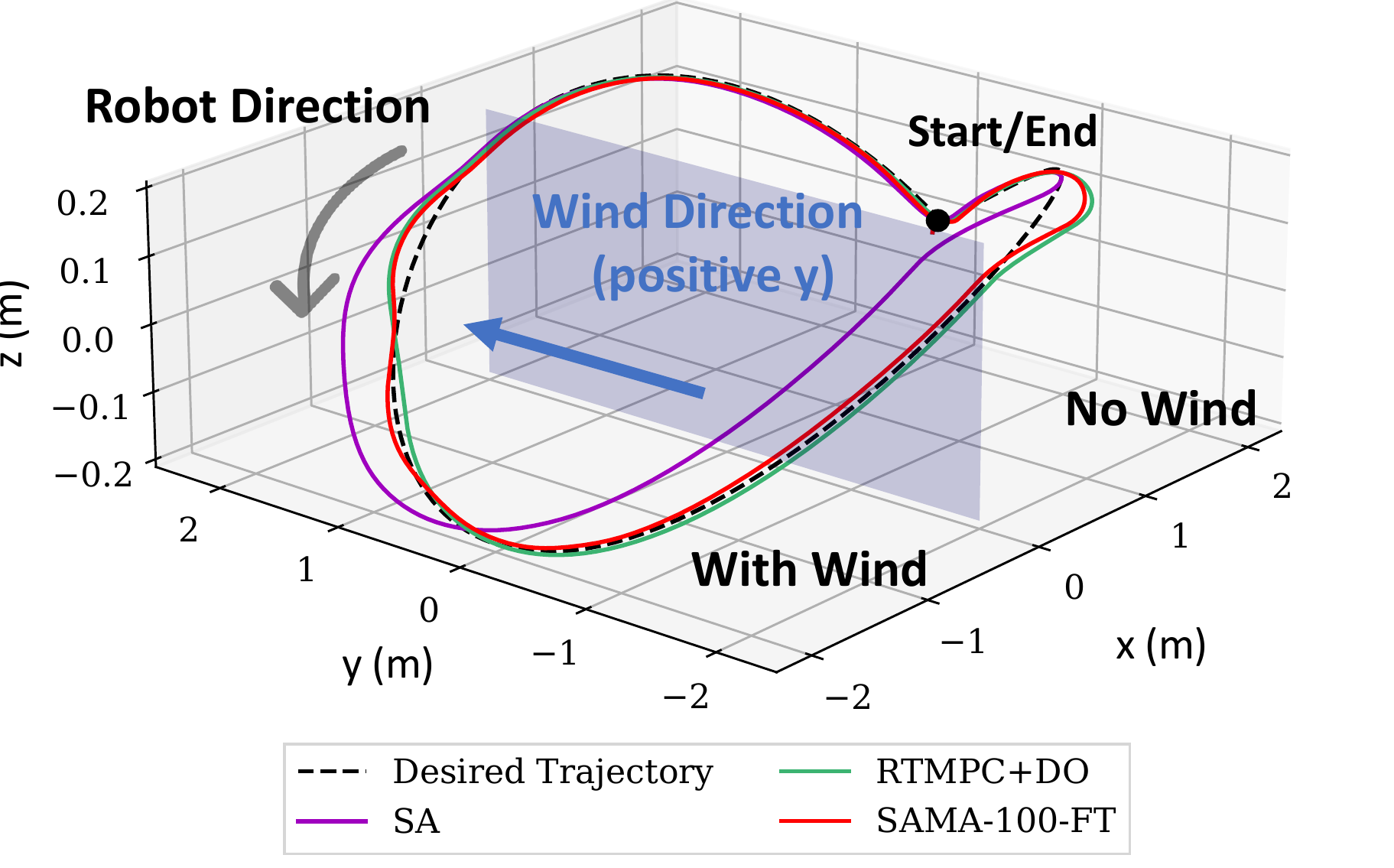}
    \caption{Tracking of a heart-shaped trajectory under strong, out-of-distribution wind-like external force disturbances. The blue plane $p_x\!=\!0$ divides the environment into a part with wind ($p_x\!\leq\!0$) and one without ($p_x\!>\!0$). Our adaptive approach (SAMA-100-FT) demonstrates an improvement on our previous work (SA), which is robust but not adaptive, and it is able to match the performance of robust MPC combined with a disturbance observer (RTMPC+DO), but at a fraction of its computational cost.}
    \label{fig:traj_combined}
    \vspace{-4ex}
\end{figure}

In this part, we analyze our approach on the task of performing \textit{Phase 0 and 1}, as these are the parts where our method introduces key changes compared to prior \ac{RMA} work, on the task of learning the heart-shaped trajectory (\cref{fig:traj_combined}). We study the performance (average position error from the reference trajectory) and robustness (avoiding violation of state and actuation constraints) of our policy as a function of the total training time and the number of demonstrations used for training. Our comparison includes the \ac{RTMPC} expert that our policy tries to imitate, as well as a policy trained only with DAgger, without any data augmentation. Each policy is evaluated on $10$ separate realizations of the training/test environment. We repeat the procedure over $6$ random seeds. All training in this part is done on a single CPU and GPU.

\begin{figure}[ht!]
    \centering\includegraphics[width=0.5\textwidth,trim = {0.2cm, 0, 0.05cm, 0}, clip]{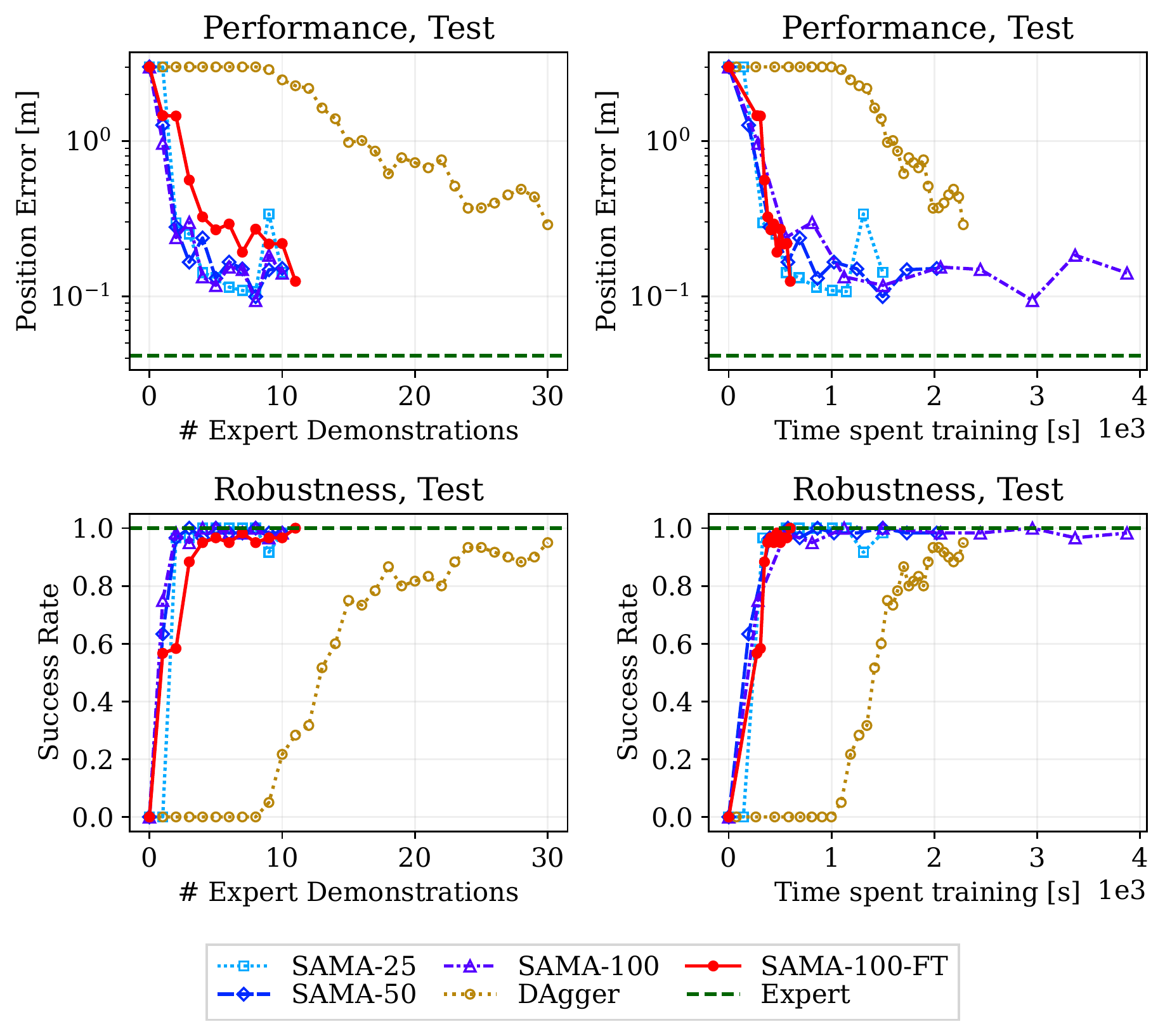}
    \captionof{figure}{Performance and robustness in the \textbf{testing environment} after \textit{Phase 1}, as a function of number of demonstrations and training time. The test environment presents a set of disturbances that the robot has never seen during training, as highlighted in \cref{table:env-params}. Methods that rely on our tube-guided data augmentation strategy (SAMA) generalize better than DAgger, achieving higher robustness and performance in lower time.}
    \label{fig:perf_robust_test}
    \vskip-2ex
\end{figure}

Figure \ref{fig:perf_robust_train} shows the evaluation of the policy under a new set of disturbances sampled from the same training distribution, defined in Table \ref{table:env-params}, highlighting that our tube-guided data augmentation strategy efficiently learns robust \textit{Phase 1} policies. Compared to DAgger-only, our methods achieve full robustness in less than \textit{half} the time, and using only 20\% of the required expert demonstrations. Additionally, tube-guided methods achieve about half the position error of DAgger for the same training time, reaching an average of $5$ cm  in less than $10$ minutes.  Among tube-guided data augmentation methods, we observe that fine-tuning (SAMA-100-FT) achieves the lowest tracking error in the shortest time.
Figure \ref{fig:perf_robust_test} repeats the evaluation under a challenging set of disturbances that are outside the training distribution (see Table \ref{table:env-params}). The analysis, as before, highlights the benefits of the data augmentation strategy, as SAMA methods achieve higher robustness and performance. The performance in this test environment confirms the trend that fine-tuning (SAMA-100-FT) achieves good trade-offs in terms of training time and robustness. Overall, these results highlight that our method can successfully and efficiently learn \textit{Phase 1} policies capable of handling out-of-distribution disturbances.

\begin{figure*}[ht!]
    \centering\includegraphics[trim=0 7 0 5, clip, width=0.825\textwidth]{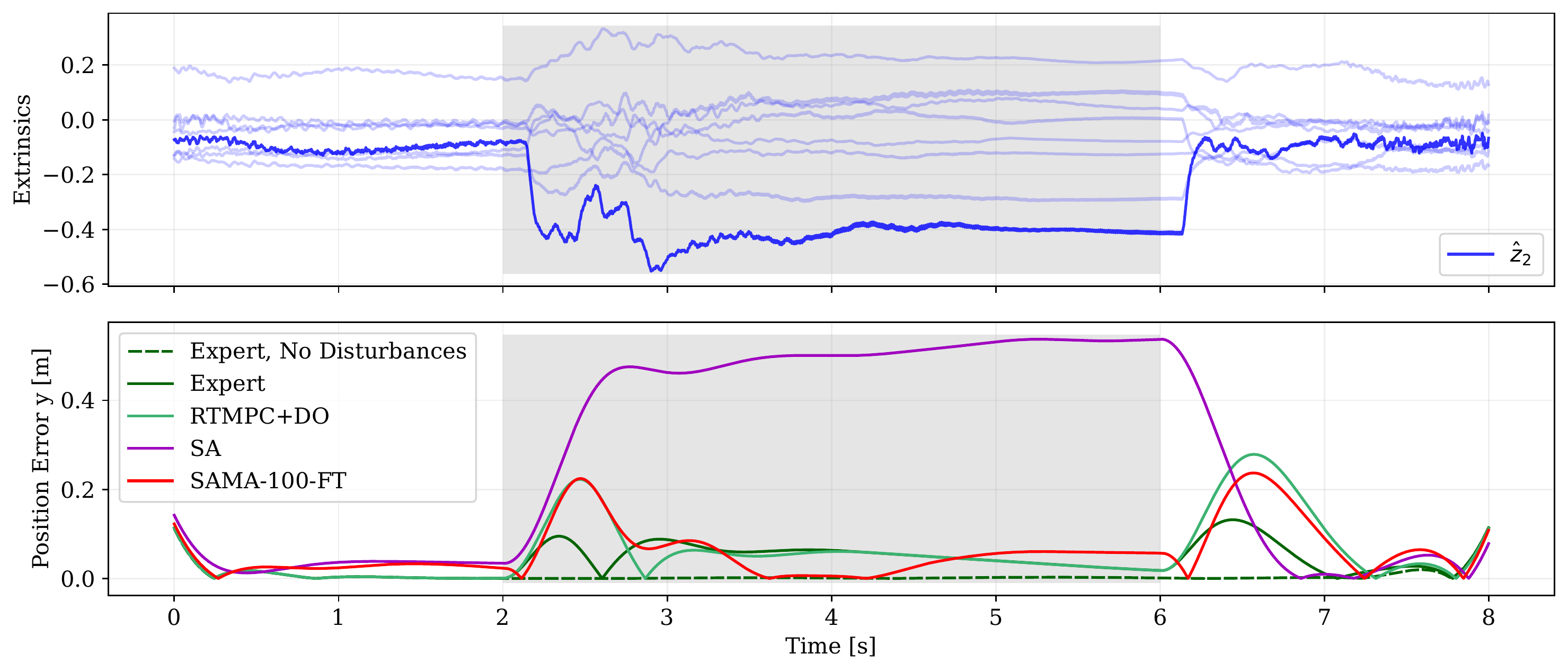}
    \caption{Performance while tracking the heart-shaped trajectory in \cref{fig:traj_combined}. The robot is subject to an out-of-training-distribution wind-like force of $6$ N (along positive $y$-axis, shown in shaded grey area) that is $36\%$ larger than any external forces seen during training. Our method (SAMA-100-FT) is computationally efficient, robust, and adaptive, as shown by the changes in extrinsics ($\hat{z}_2$) when the robot is subject to wind. Our method achieves $10\%$ lower tracking error than RTMPC+DO, a model-based controller which is both robust and adaptive at the cost of being computationally expensive to run online (see Table \ref{table:runtime}), and which has been designed using a nominal model. We additionally maintain similar performance to the Expert, an RTMPC that has access to the ground truth model and disturbances and represents the best case performance of a model-based controller. This highlights improvements over our previous work SA \cite{tagliabue2022demonstration}, a learning-based controller which is robust and computationally efficient, but non-adaptive.}
    \label{fig:traj_and_extrinsics}
    \vspace{-3ex}
\end{figure*}

\subsection{Adaptation Performance Evaluation}
In this part, we analyze the adaptation performance of our approach after \textit{Phase 2}. We consider the heart-shaped and the eight-shaped trajectory. For each trajectory, we train a $\mu$ and $\pi$ in \textit{Phase 1} using SAMA-100-FT, fine-tuning for $5$ DAgger iterations, collecting $10$ demonstrations per iteration during fine-tuning. Given a trained $\mu$ and $\pi$, we train $\phi$ via supervised regression in \textit{Phase 2} (Section \ref{training-details}), conducting $20$ iterations with $10$ policy rollouts collected in parallel per iteration. On $10$ CPUs and $1$ GPU, \textit{Phase 1} takes about $20$ minutes and \textit{Phase 2} takes about an hour. We note that the training efficiency of our proposed \textit{Phase 1} compares favorably to the \ac{RL}-based results in \cite{zhang2022zero}, where the authors report $2$ hours of training time for \textit{Phase 1}. 

\begin{figure}
    % trim={<left> <lower> <right> <upper>}
    \centering\includegraphics[trim=380 95 100 105,clip,width=0.867\columnwidth]{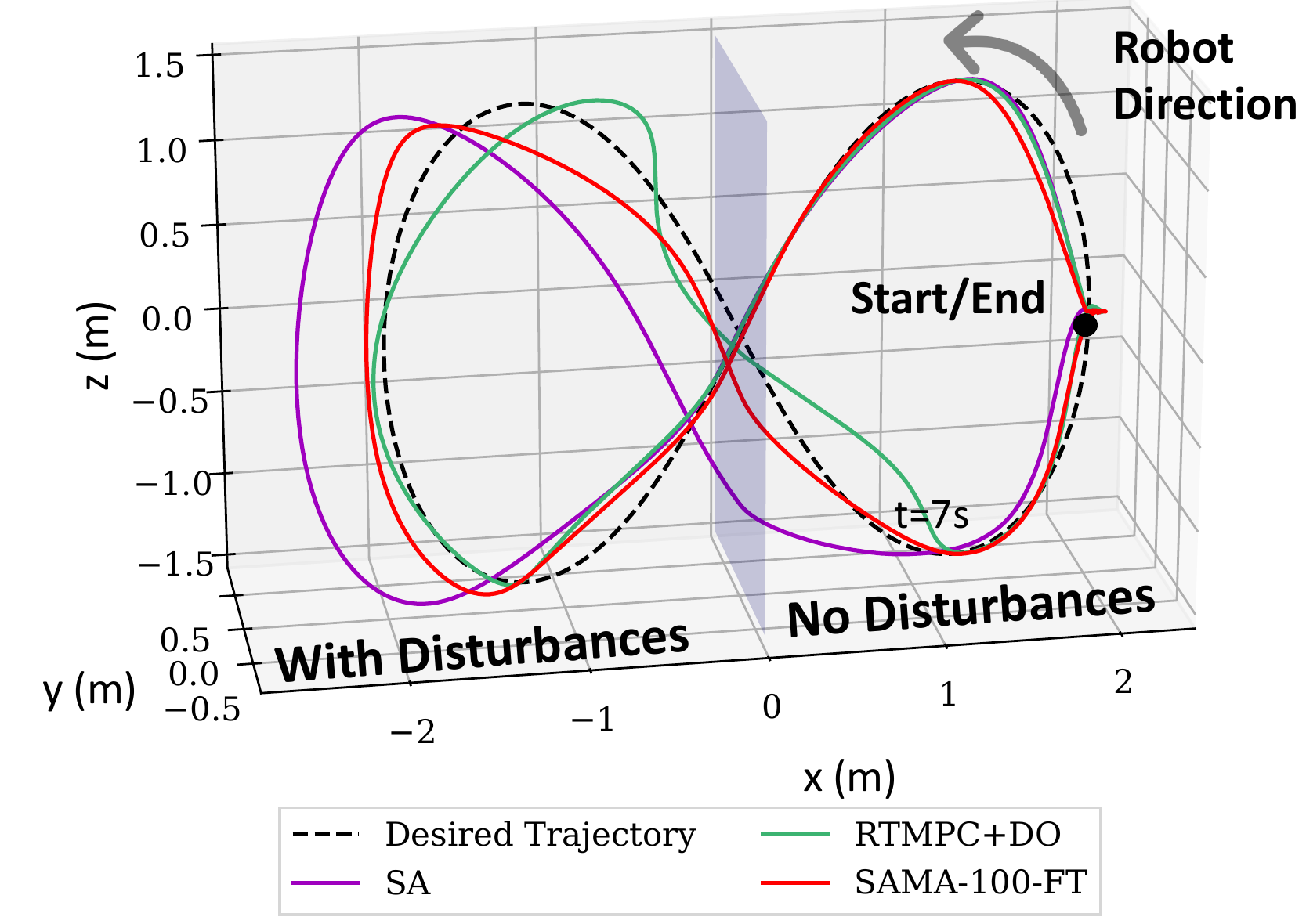}
    \caption{Tracking of an eight-shaped trajectory under out-of-distribution disturbances and model errors, where mass and arm length are twice the nominal values, the drag is $10$ times the nominal, and there is a $2$\,Nm external torque disturbance. The blue plane $p_x\!=\!0$ divides the environment into a part with model errors ($p_x\!\leq\!0$) and without ($p_x\!>\!0$). Our adaptive approach (SAMA-100-FT) can adapt during agile flight, reaching top speeds of $3.2$\,m/s, while maintaining performance comparable to RTMPC+DO.}
    \label{fig:traj_combined_lem}
    \vspace{-3ex}
\end{figure}

First, we evaluate the tracking performance of our adaptive controller in an environment subject to position-dependent winds, as shown in \cref{fig:traj_combined}. The wind applies $6$ N of force, a force $36\%$ \textit{larger} than any external force encountered during training. We compare our approach with \ac{SA}, our previous non-adaptive robust policy learning method \cite{tagliabue2022demonstration}, and with the \ac{RTMPC} expert that has access to $e_t$ (the true value of the wind). We also consider an \ac{RTMPC} whose state has been augmented with external force/torques estimated via a state-of-the-art nonlinear disturbance observer (\ac{RTMPC}+\ac{DO}) based on an \ac{UKF} \cite{tagliabue2019robust}, a method that has access to the nominal model of the robot (matching the one used in this experiment) and ad-hoc external force/torque disturbance estimation. The results are presented in \cref{fig:traj_and_extrinsics} and \cref{fig:traj_combined}. The shaded section of \cref{fig:traj_and_extrinsics}, corresponding to the windy regions,  highlights that \ac{SAMA}-100-FT is able to adapt to a large, previously unseen force-like disturbance, obtaining a tracking error of less than $10$ cm at convergence, unlike the corresponding non-adaptive variant (\ac{SA}), which instead suffers from a $50$ cm tracking error. \cref{fig:traj_and_extrinsics} additionally highlights changes in the extrinsics, which do not depend on changes in reference trajectory but rather on the presence of the wind, confirming the successful adaptation of the policy. \cref{tab:tracking_performance} reports a $10 \%$ reduction in tracking error compared to \ac{RTMPC}+\ac{DO}. Second, we repeat the evaluation on the challenging eight-shaped trajectory, with the robot achieving speeds of up to $3.2$ m/s, where the robot is subject to a large set of out-of-distribution model errors: twice the nominal mass and arm length, ten times the nominal drag coefficients, and an external torque of $2.0$\,Nm. 
\cref{tab:tracking_performance} and \cref{fig:traj_combined_lem} highlight the adaptation capabilities of our approach, which performs comparably to the more computationally expensive (\cref{table:runtime}) \ac{RTMPC}+\ac{DO}.

\noindent
\textbf{Efficiency at Deployment.} Table \ref{table:runtime} reports the time to compute a new action for each method. On average, our method (\ac{SAMA}-100-FT) is $12$ times faster than the expert, and $24$ times faster than \ac{RTMPC}+\ac{DO}.

\begin{table}[t] 
\caption{Average Position Error (APE) while tracking  $8$\,s-long trajectories $\heartsuit$ and $\infty$ in the test environment. Each policy was evaluated over $30$ realizations of the test environment. Our approach (SAMA-100-FT) achieves successful adaptation, obtaining lower error than our previous, non-adaptive strategy (SA), and lower or comparable errors  to RTMPC+DO.}
\vspace{-1em}
\begin{center}
\begin{tabular}{|r|p{.25\columnwidth}|p{.25\columnwidth}|}
\hline
\textbf{Method}                                       & \textbf{APE} $\heartsuit$ [m]           & \textbf{APE} $\infty$ [m]              \\ \hline \hline
Expert                                                & \hfil0.058                              & \hfil0.104                             \\ 
RTMPC+DO                                              & \hfil0.125                              & \hfil0.163                             \\ 
SA (not adaptive) \cite{tagliabue2022demonstration}   & \hfil0.278                              & \hfil0.322                             \\ 
SAMA-100-FT (Ours)                            & \hfil0.110                              & \hfil0.175                             \\  \hline
\end{tabular}
\end{center}
\label{tab:tracking_performance}
\vskip-2ex
\end{table}

\begin{table}[t] 
\caption{Time required to generate a new action; all times reported in milliseconds (ms). Our approach (SAMA-100-FT) is on average $12\times$ faster than the optimization-based Expert, and $24\times$ faster than an optimization-based approach with disturbance observer (RTMPC+DO). While our previous work (SA) \cite{tagliabue2022demonstration} achieves a faster inference time than our method, it lacks adaptation, which our method adds with minimal computational cost.} \label{table:runtime}
\vspace{-0.3em}
\begin{center}
\begin{tabular}{
|@{\hspace{0.5em}}r@{\hspace{0.5em}}
|@{\hspace{0.5em}}c@{\hspace{0.5em}}
|@{\hspace{0.5em}}c@{\hspace{0.5em}}
|@{\hspace{0.5em}}c@{\hspace{0.5em}}
|@{\hspace{0.5em}}c@{\hspace{0.5em}}
|@{\hspace{0.5em}}c@{\hspace{0.5em}}|}
\hline
\textbf{Method} & \textbf{Setup} & \textbf{Mean} & \textbf{SD} & \textbf{Min} & \textbf{Max} \\ \hline \hline
Expert                                       & CVXPY     & 9.51 & 6.16 & 5.04 & 62.2 \\ 
RTMPC+DO                                     & CVXPY     & 19.0 & 14.0 & 13.5 & 937 \\ 
SA \cite{tagliabue2022demonstration}         & PyTorch   & 0.491 & 7.55e-2 & 0.458 & 1.54 \\ 
SAMA-100-FT (Ours)                   & PyTorch   & 0.772 & 3.68e-2 & 0.669 & 1.58 \\  \hline
\end{tabular}
\end{center}
\vskip-3ex
\end{table}

\section{Conclusions} \label{sec:conclusions}
We presented \ac{SAMA}, a strategy to enable adaptation in policies efficiently learned from a Robust Tube MPC expert using Imitation Learning and \ac{SA} \cite{tagliabue2022demonstration}, a tube-guided data augmentation strategy. We did so by leveraging an adaptation scheme inspired by the recent Rapid Motion Adaptation work \cite{kumar2021rma}. Our evaluation in simulation has demonstrated successful learning of an adaptive, robust policy that can handle strong out-of-training distribution disturbances while controlling the position and attitude of a multirotor. Future work will experimentally evaluate the approach and perform a more direct comparison with \ac{RMA} for quadrotors \cite{zhang2022zero}, once code becomes available.

%\section*{ACKNOWLEDGMENT}
%This work was funded by the Air Force Office of Scientific Research MURI FA9550-19-1-0386. We thank Lauren Li, Xiaoyi (Jeremy) Cai, Kota Kondo and Yulun Tian for reviewing the manuscript.  

\balance
\bibliographystyle{IEEEtran}
\bibliography{bibliography}

\end{document}